\documentclass[conference]{IEEEtran}
\IEEEoverridecommandlockouts
\usepackage{cite}
\usepackage{geometry}
\usepackage{amsmath,amssymb,amsfonts}
\usepackage{algorithmic}
\usepackage{afterpage}
\usepackage{graphicx,booktabs, multirow}
\usepackage{textcomp}
\usepackage{xcolor}
\usepackage{float}
\usepackage{makecell}
\usepackage{array}
\def\BibTeX{{\rm B\kern-.05em{\sc i\kern-.025em b}\kern-.08em
    T\kern-.1667em\lower.7ex\hbox{E}\kern-.125emX}}

\begin{document}

\newgeometry{left=0.75in, right=0.75in, top=1in, bottom=0.75in}

\title{Speech as a Multimodal Digital Phenotype for Multi-Task LLM-based Mental Health Prediction\\

}
\author{
    \IEEEauthorblockN{Mai
    Ali\textsuperscript{1}, Christopher Lucasius\textsuperscript{*1}, Tanmay P. Patel\textsuperscript{*2}, Madison Aitken\textsuperscript{3,4}, Jacob Vorstman\textsuperscript{5,6}, Peter Szatmari\textsuperscript{3,5,6}, \\ 
    Marco Battaglia\textsuperscript{6}, Deepa Kundur\textsuperscript{1}} 
    \IEEEauthorblockA{\textsuperscript{1}The Edward S. Rogers Sr. Department of Electrical and Computer Engineering, University of Toronto, Toronto, Canada \\  
    \textsuperscript{2}Division of Engineering Science, University of Toronto, Toronto, Canada \\  
    \textsuperscript{3}Cundill Centre for Child and Youth Depression, Centre for Addiction and Mental Health, Toronto, Canada \\  
    \textsuperscript{4}Department of Psychology, York University, Toronto, Canada \\  
    \textsuperscript{5}The Hospital for Sick Children, Toronto, ON, Canada \\  
    \textsuperscript{6}Department of Psychiatry, University of Toronto, Toronto, Canada
    }  
    Emails: \{maia.ali, christopher.lucasius, tanmay.patel\}@mail.utoronto.ca \\
    aitken@yorku.ca, jacob.vorstman@sickkids.ca, \{peter.szatmari, marco.battaglia\}@utoronto.ca, dkundur@ece.utoronto.ca
}

\maketitle
\def\thefootnote{*}\footnotetext{These authors contributed equally to this work.}\def\thefootnote{\arabic{footnote}}

\begin{abstract}
Speech is a noninvasive digital phenotype that can offer valuable insights into mental health conditions, but it is often treated as a single modality. In contrast, we propose the treatment of patient speech data as a trimodal multimedia data source for depression detection. This study explores the potential of large language model-based architectures for speech-based depression prediction in a multimodal regime that integrates speech-derived text, acoustic landmarks, and vocal biomarkers. Adolescent depression presents a significant challenge and is often comorbid with multiple disorders, such as suicidal ideation and sleep disturbances. This presents an additional opportunity to integrate multi-task learning (MTL) into our study by simultaneously predicting depression, suicidal ideation, and sleep disturbances using the multimodal formulation. We also propose a longitudinal analysis strategy that models temporal changes across multiple clinical interactions, allowing for a comprehensive understanding of the conditions' progression. Our proposed approach, featuring trimodal, longitudinal MTL is evaluated on the Depression Early Warning dataset. It achieves a balanced accuracy of 70.8\%, which is higher than each of the unimodal, single-task, and non-longitudinal methods.

\end{abstract}


\section{Introduction}
Depression, suicidal ideation, and sleep disturbances are prevalent and interconnected mental health problems among adolescents. Major depressive disorder (MDD) affects 8–12\% of adolescents globally \cite{lu_2019_adolescent}, while suicide is the second leading cause of death among individuals aged 15–24 \cite{lu_2019_adolescent}. Sleep disturbances impact 20–50\% of adolescents, increasing the risk of depression and suicidality \cite{gradisar_2022_sleeps}.

Depression typically emerges between the ages of 12 and 18, with 20\% of adolescents experiencing a major depressive episode by 18 \cite{inkelis_2020_elevated}. Sleep disturbances, driven by biological and lifestyle factors, are strongly linked to depression and suicidal ideation. Notably, sleep deprivation independently raises the risk of suicidality, with each hour of sleep loss increasing suicidal thoughts by 11\%\cite{inkelis_2020_elevated}. Given the bidirectional relationships among these conditions, integrating sleep and suicidality assessments into depression screenings could enhance early detection and intervention.

\subsection{The Need for Multimodal AI-Driven Mental Healthcare}
Speech provides a non-invasive and suprisingly multimodal source for assessing mental health, allowing for the extraction of linguistic (word choice, syntax) and paralinguistic (tone, pitch, rhythm) features \cite{corcoran_2020_language}. Studies have shown that individuals with depression and suicidal ideation exhibit distinct speech characteristics, including slower speech rate, longer pauses, and increased monotonicity \cite{KAPPEN2023105121}. Additionally, linguistic patterns in speech, such as self-referential language and negative emotion words, have been strongly linked to depression and suicidality \cite{odea_2021_the}.

Recent advancements in large language models (LLMs) have revolutionized speech and text-based mental health assessments. LLMs pre-trained on large corpora of conversational and clinical data capture subtle linguistic and acoustic markers associated with depression, suicidality, and sleep disturbances, making them well-suited for automated screening in adolescents \cite{Xu_2024}. In addition, multi-task learning (MTL) approaches allow LLMs to model shared representations across multiple mental health conditions, improving generalization and reducing the need for extensive labeled data \cite{berrenursaylam_2024_multitask}.

\subsection{Contributions}
\label{sec:contribution}
While the use of LLMs for depression detection has grown into a rich field, there are several notable gaps. We offer the following innovations in this space:
\begin{enumerate}
    \item A comprehensive multimedia framework for depression detection that integrates three speech-derived modalities: speech transcriptions, acoustic landmarks \cite{Zhang_2024}, and vocal biomarkers. This unified approach provides a 
\end{enumerate}
\clearpage
\restoregeometry  

\begin{enumerate}
    \setcounter{enumi}{1}  
    \item[] more holistic understanding of speech-based depression indicators.
    \item A longitudinal analysis framework that tracks changes across multiple patient-physician interactions by treating the interactions collectively as a continuous LLM `conversation'. This type of analysis is not novel; however, to the best of our knowledge, we are the first to apply it to multimodal speech and text-based mental health analysis, an area where temporal understanding is typically viewed by clinicians to be critical \cite{obradovich_2024_opportunities}.
    \item A multi-task learning architecture that extends our trimodal approach beyond depression detection to related clinical research assessments, leveraging multimodal speech data and maximizing the utility of data collected across multiple related domains.
\end{enumerate}

We assess these contributions relative to current state-of-the-art methods on the Depression Early Warning (DEW) dataset, a longitudinal multimodal dataset described in Section \ref{sec:dataset}.

\section{Background and Literature Review}
The potential of machine learning models for predicting mental health conditions has been well established in the literature. Prior studies have explored various modalities as predictive biomarkers, including actigraphy-based activity and sleep patterns, ecological momentary assessments, facial expressions, and speech characteristics \cite{lucasius_prediction}. Recent advancements in LLMs have enhanced their ability to process long-form transcripts and infer underlying cognitive and emotional states. These models present a novel opportunity to extract clinically relevant indicators of mental health conditions from speech data, facilitating more accurate and scalable diagnostic support systems \cite{stade_2024_large}.

\subsection{Prediction Based on Speech Semantics and Text}
\label{sec:lit-review-text}
In recent months, the use of LLMs to detect depression from patients' text or transcribed speech data has become an expanding research area. A central work is that of Xu et al. \cite{Xu_2024} which performs systematic benchmarking of widespread general-purpose LLMs against mental health classification tasks. They find limited, but promising, potential in zero-shot and few-shot regimes. Their work introduces two new LLMs---Mental-Alpaca and Mental-FLAN-T5---that were instruction fine-tuned for multi-task mental health classification. They outperform significantly larger pre-trained models \cite{Xu_2024}.

Another active research area is the use of longitudinal methods for LLM-based depression detection that leverage connections between text samples from different points in a patient's history. They capture a patient's treatment trajectory and use the conversational nature of LLMs to analyze this dimension. However, existing works employ longitudinal methods in single modality regimes, chiefly on text data. One example is \cite{Qin_2023_LLMSocialMedia} which uses tweets collected over long periods to engage an LLM in a `conversation' that generates a prediction capable of capturing the time-dependence of patient data. We extend these ideas to our multimodal framework by treating each clinical research assessment as an episode that's part of a larger LLM interaction (see Section \ref{sec:proposed-pipeline-final}).

\subsection{Prediction Based on Vocal Biomarkers}
Several studies have explored the use of various speech features, with an emphasis on vocal biomarkers (e.g., fundamental frequency, mel-frequency cepstrum coefficients (MFCC)), which have been found to correlate with mental disorders. Tasnim et al. propose a dataset for machine learning-based depression detection that extracts hand-curated vocal features from patient speech samples (based on clinical domain knowledge) to enhance predictive power \cite{Tasnim_2023}. Our work unifies these biomarkers with other speech-based modalities, discussed below, that are amenable to LLM-based analysis. The biomarkers we use are outlined in Section \ref{sec:audio-features}.

\subsection{Prediction Based on Acoustic Landmarks}
Although transcribed speech data serve as a natural input to language models, speech data contain other multimedia features that are amenable to LLM-based analysis for depression detection. An approach introduced in \cite{Zhang_2024} extracts acoustic landmarks from speech samples. These landmarks are a discretized sequence of symbols that represent linguistic and pronunciation patterns; they add a critical dimension to the raw transcripts \cite{Zhang_2024}. Their work follows a two-stage strategy to take advantage of these landmarks alongside the text: (a) they fine-tune the LLM with Low-Rank Adaptation matrices (LoRA) \cite{Hu_2021_LORA} to develop a latent representation of the landmarks; (b) they attach a classifier to the LLM and employ prompt (P)-tuning on the combined model for depression classification. They achieve state-of-the-art results with their multimodal approach. A notable omission, however, is the lack of features from the speech waveform itself, such as vocal biomarkers. Such modalities are less compliant with LLM-based methods, but are a key component of our proposed architecture.

\subsection{Multi-Task Learning for Mental Heath Prediction}
Multi-task learning is a machine learning paradigm in which a single model is trained to perform multiple related tasks simultaneously by sharing common representations. This approach allows the model to learn underlying patterns that are shared among tasks, often leading to better generalization and robustness compared to models trained on individual tasks. In essence, MTL leverages inter-task relationships to improve performance, reduce overfitting, and efficiently utilize limited data \cite{caruana_1997_multitask}.
Recent research has explored MTL for mental health classification using various data sources. In \cite{benton_2017_multitask}, Benton et al. demonstrated the effectiveness of MTL in predicting mental health conditions from social media text, particularly for conditions with limited data. They showed that combining demographic attributes and mental states in an MTL framework outperformed single-task models. Azim et al. applied  MTL with a Bi-LSTM to detect changes in mood and suicidal risk levels in longitudinal user texts, outperforming single-task frameworks \cite{azim_2022_detecting}.

\section{Problem Formulation}
In this section, we present the Depression Early Warning (DEW) dataset utilized in this study. We provide a description of the collected modalities, the data acquisition protocol, and the pre-processing steps applied to the raw patient data to ensure suitability for analysis. Subsequently, we outline the methodology for constructing task labels corresponding to depression, suicidal ideation, and sleep disturbances. Lastly, we introduce the experimental design for this analysis.



\subsection{The Depression Early Warning (DEW) Dataset}
\label{sec:dataset}
The DEW project is conducted at the Centre for Addiction and Mental Health (CAMH) in Canada. The study is focused on gaining insights into the behavior of adolescents with a history of depression, specifically Major Depressive Disorder (MDD), within the age range of 12-21 years. 


The multimodal speech samples used in this analysis are collected during interviews that are conducted every follow-up visit (\texttt{arm}). The samples contain responses from participants to questions while interviewed by the DEW study research coordinators, usually expressing their opinions on various topics. Hence, each participant, designated by \texttt{patient\_id}, contributes a single speech sample per follow-up session (\texttt{arm}). Our experimental setup proposes two architecture variations: a longitudinal approach that considers a participant's previous \texttt{arms} to predict their current mental health state, and a cross-sectional approach treats each \texttt{arm} independently. Further details on these approaches are provided in Section \ref{sec:proposed_method}.


\subsection{Task Labels}
This study involves three binary classification tasks derived from participant surveys completed at follow-up visits. The primary task is depression, labeled using either a score from the Children’s Depression Rating Scale (CDRS) or the Hamilton Depression Rating Scale (HAM-D), which are unified via an equipercentile method for consistency \cite{Furukawa_2019}.

The first auxiliary task, suicidal ideation, is identified through responses to question 9 of the Patient Health Questionnaire (PHQ-9) and question 19 of the Mood and Feelings Questionnaire (MFQ). A participant is labeled as exhibiting suicidal ideation if they score 1 on both scales. The validity of these survey questions as screeners for suicidal ideation has been described in \cite{simon_2013_does} and \cite{hammerton_2014_validation} respectively. 

The second auxiliary task, sleep disturbances, is determined using question 3 of the PHQ-9 and questions 32 and 33 of the MFQ \cite{macgregor_2012_evaluation}. Further details on survey items, scoring criteria, and labeling thresholds are outlined in Table~\ref{tab:task_labels}.




\begin{table*}[ht] 
    \caption{Depression, Suicidal Ideation, and Sleep Disturbances Tasks Labeling}
    \label{tab:task_labels}
    \centering
    \renewcommand{\arraystretch}{1.2} 
    \setlength{\tabcolsep}{4pt} 
    \begin{tabular}{|p{2.5cm}|p{5cm}|p{3cm}|p{3.5cm}|p{0.7cm}|}
        \hline
        \textbf{Task} & \textbf{Survey/Question} & \textbf{Scale} & \textbf{Cut-off} & \textbf{Label} \\
        \hline
        Depression & HAM-D & 10 - 13: mild. \newline 14-17: mild to moderate. \newline \textgreater 17: moderate to severe. & \textgreater 17 \newline \textless 17  & 0 \newline 1 \\
        \hline
        Suicidal Ideation & PHQ-9/ Q9 (Thoughts that you would be better off dead, or of hurting yourself) & 0: not at all \newline 3: nearly everyday & PHQ-9/Q9 + MFQ/Q19 \textgreater 2  & 1 \\
        & MFQ/Q19 (I thought about killing myself) & 0: not true \newline 2: true &  PHQ-9/Q9 + MFQ/Q19 \textless 2  & 0 \\
        \hline
        Sleep Disturbances & PHQ-9/ Q3 (Trouble falling or staying asleep, or sleeping too much) & 0: not at all \newline 3: nearly everyday & PHQ-9/ Q9 + MFQ/ Q32 + MFQ/ Q32  \textgreater 3 & 1 \\
        & MFQ/ Q32 (I didn’t sleep as well as I usually sleep.) & 0: not true \newline 2: true &  &  \\
        & MFQ/ Q33 (I slept a lot more than usual) & 0: not true \newline 2: true & PHQ-9/ Q9 + MFQ/ Q32 + MFQ/ Q32  \textless 3 & 0 \\
        \hline
    \end{tabular}
\end{table*}

\subsection{Experimental Design}
We employ binary classification for the three conditions and use three distinct architectures based on the following sets of modalities: text; text \& acoustic landmarks; and text, acoustic landmarks \& vocal biomarkers. Each architecture is designed for a particular modality set and is built on top of two `LLM bases'---the general-purpose LLaMA-2-7B model \cite{Houvron_2023_Llama} and the mental health-specific Mental-Alpaca model. This enables a comparison across modality sets, as well as between the two LLMs. The goal is to develop and evaluate a holistic multimodal, multi-task framework that can predict these conditions simultaneously, utilizing metrics such as precision, recall, and balanced accuracy. The experimental setup focuses on minimizing a combined loss function across the three tasks to improve cross-modality generalization and robustness.
 
\section{Proposed Method}
\label{sec:proposed_method}
We begin by describing the extraction and tokenization processes for the three modalities leveraged in this work: text, acoustic landmarks, and vocal biomarkers. We then describe the three model architectures and training pipelines.

\subsection{Feature Extraction and Tokenization}
The proposed architecture treats the patients' speech data as a trimodal multimedia source composed of text, acoustic landmarks, and audio biomarker features. This section details the extraction of these three components.

\subsubsection{Text}
Text transcripts of the speech data are generated using OpenAI's Whisper Speech Recognition System. While this software is known to be robust \cite{radford2022robustspeechrecognitionlargescale}, we manually scanned the generated transcripts and looked for incomplete sentences. We manually transcribed these sentences. Although this method is not flawless, it is feasible and still provides significant improvements over blind AI-based transcription.

\subsubsection{Landmarks}
Acoustic landmarks are extracted as per the procedure proposed by \cite{Boyce_2012} and detailed in \cite{Zhang_2024}. The audio spectrogram of each patient is divided into six frequency bands. Energy changes in one or more of these bands are classified under various landmark symbols. Some examples of landmarks include the vibration of vocal folds, the release or closure of the nasal passage, voiced frication, periodicity, etc. The sequence of landmarks corresponding to a specific speech sample is recorded alongside the text transcriptions for consumption by the proposed architectures.

\subsubsection{Vocal Biomarkers}
\label{sec:audio-features}
Vocal biomarkers are extracted using Python's Librosa library in accordance with \cite{Tasnim_2023}. We divide each speech sample into 500-millisecond windows and extract a set of summary statistics from each window. These features encompass spectral characteristics such as sound intensity, MFCC, delta-MFCC, pitch, magnitude, and zero-crossing rate (ZCR), along with voicing-related attributes, including fundamental frequency ($F_0$), harmonicity, harmonic-to-noise ratio (HNR), shimmer and jitter, energy, durational features, pauses, fillers, and phonation rate. These features were used by Tasnim et al. \cite{Tasnim_2023} to achieve state-of-the-art performance on the Audio Visual Emotion Challenge depression dataset \cite{Ringeval_2019}.

\begin{figure*}[htbp]
    \centering
    \includegraphics[width=0.92\textwidth, trim=0 8cm 0 8cm, clip]{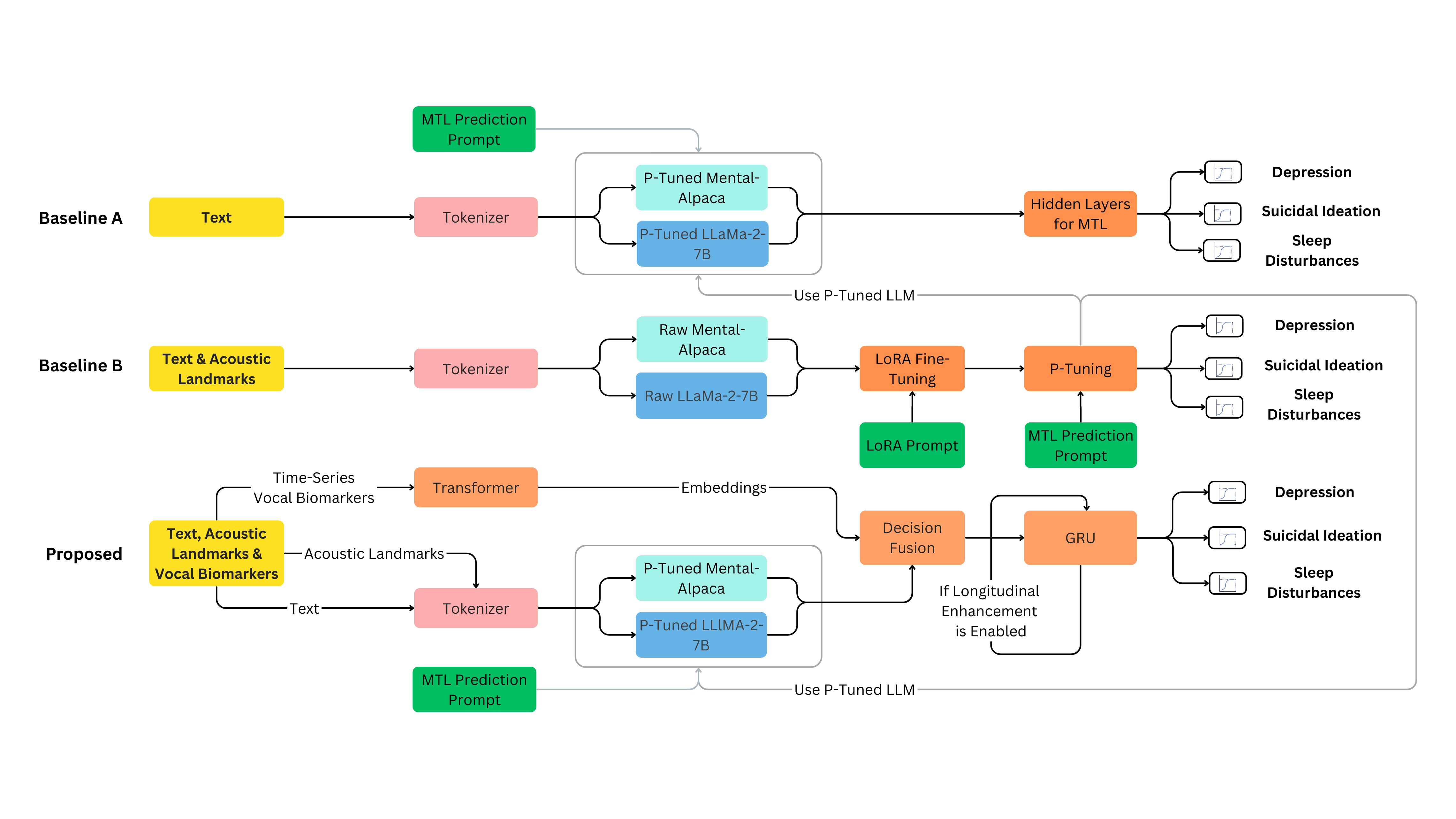} 
    \caption{A summary of the three architectures and corresponding pipelines studied in this paper.} 
    \label{fig:pipeline}
\end{figure*}

\subsection{Baseline A: Text Model with Multi-Task Learning}
This baseline employs a P-tuned version of either Mental-Alpaca (see Section \ref{sec:lit-review-text}), or Meta AI's LLaMA-2-7B \cite{Houvron_2023_Llama}. Both models have potential since Mental-Alpaca is fine-tuned to mental health tasks, while LLaMA-2-7B has shown promise on general-purpose prediction; this justifies a comparison. The LLM prediction (represented as an embedding) is then used for MTL, as outlined in Section \ref{sec:mtl-all-models}.

\subsection{Baseline B: Text and Acoustic Landmark Pipeline with Multi-Task Learning}
\label{sec:architecture}
This pipeline is heavily influenced by the work of Hu et al. \cite{Zhang_2024}. We replicate their two-stage procedure: hint cross-modal instruction fine-tuning, followed by P-tuning for depression detection \cite{liu2023gptunderstands}. Similar to baseline A, we extend the work to support MTL and allow support for both LLaMA-2-7B and Mental-Alpaca.

In cross-modal fine-tuning, an LLM is provided with prompts that provide instructions to map the corresponding acoustic landmarks given a transcript. This allows the LLM to learn the semantics of acoustic landmarks and align the positions of text to landmarks. Since the LLM has billions of trainable parameters, we use the LoRA technique \cite{Xu_2024} to only train a limited number of parameters. After the LLM is trained to recognize the combined text and landmark data, a different prompt is fed into the LLM to ask it to predict three binary labels: depression, suicidal ideation, and sleep disturbances. Instead of using the usual language model head, a fully connected classification layer is added to the LLM to make these predictions. P-tuning is applied to add trainable prompt embeddings in combination with the original prompt to fine-tune the LLM during training.

\begin{table*}[ht]
    \centering
    \renewcommand{\arraystretch}{1.2}
    \setlength{\tabcolsep}{6pt}
    \caption{Multi-task Results Across Architectures/Modalities and Base LLMs; P = precision, R = recall, BA = balanced accuracy}
    \label{tab:main-results}
    \begin{tabular}{|c|c|ccc|ccc|}
        \hline
        \multirow{2}{*}{\textbf{Architecture/Modalities}} & \multirow{2}{*}{\textbf{Task}} & \multicolumn{3}{c|}{\textbf{Mental-Alpaca}} & \multicolumn{3}{c|}{\textbf{LLaMA-2-7B}} \\  
        \cline{3-8}
        & & \textbf{P} & \textbf{R} & \textbf{BA} & \textbf{P} & \textbf{R} & \textbf{BA} \\  
        \hline
        \multirow{3}{*}{\textbf{Baseline}: Text} & \textbf{Main}: Depression &0.362 &0.607 &0.606 &0.258 &0.537 &0.524 \\  
        & Suicidal Ideation & 0.000 & 0.000 &0.500 & 0.000 &0.000 & 0.500\\  
        & Sleep Disturbances &0.885 &1.00 &0.500 & 0.885 & 1.000 & 0.500 \\  
        \hline
        \multirow{3}{*}{\makecell{\textbf{Baseline}: Text \& Acoustic \\ Landmarks}} & \textbf{Main}: Depression & 0.291 & 0.460& 0.518& 0.302& 0.520& 0.533\\  
        & Suicidal Ideation & 0.215& 0.472& 0.524& 0.219& 0.694& 0.542\\  
        & Sleep Disturbances & 0.929& 0.404& 0.583& 0.898& 0.273& 0.518\\  
        \hline
        \multirow{3}{*}{\makecell{\textbf{Proposed}: Text, Acoustic \\ Landmarks \& Vocal Biomarkers}} & \textbf{Main}: Depression & 0.402 & 0.660 & 0.644 & 0.185 & 0.300 & 0.400 \\  
        & Suicidal Ideation & 0.727 & 0.222 & \textbf{0.601} & 0.201 & 0.778 & 0.509 \\  
        & Sleep Disturbances & 1.000 & 0.099 & 0.550 & 0.885 & 1.000 & 0.500 \\  
        \hline
        \multirow{3}{*}{\makecell{\textbf{Proposed}: Text, Acoustic \\ Landmarks \& Vocal Biomarkers \\ with Longitudinal Enhancement}} & \textbf{Main}: Depression & 0.425 & 0.680 & \textbf{0.666}& 0.271 & 0.580 & 0.495 \\  
        & Suicidal Ideation & 0.235 & 0.639 & 0.563 & 0.286 & 0.056 & 0.511 \\  
        & Sleep Disturbances & 0.924 & 0.752 & \textbf{0.638}& 0.957 & 0.280 & 0.592 \\  
        \hline
    \end{tabular}
\end{table*}

\subsection{Proposed Pipeline: Text, Acoustic Landmark, and Vocal Biomarkers for Longitudinal and Multi-Task Learning}
\label{sec:proposed-pipeline-final}

This architecture unifies all three speech-derived modalities into a novel, multimodal depression detection system. It also introduces an additional dimension of time-awareness by supporting longitudinal analysis across multiple clinical visits. A high-level representation of the architecture, along with the baselines, is provided in Fig. \ref{fig:pipeline}.

The model uses the result of baseline B---a P-tuned LLM---to generate final embeddings from the text data and their corresponding acoustic landmarks. Since this model has been pre-trained to analyze text and acoustic landmark samples for depression, its weights can remain frozen.

To analyze the vocal biomarkers extracted from the speech samples, we generate contextualized embeddings through a transformer encoder that captures the time-series nature of the biomarkers. The two embeddings are fused at the decision-level with trainable weights which are then used for MTL.

This architecture allows the user to optionally track a hidden latent vector between subsequent patient visits. This hidden layer is propagated from visit to visit through a Gated Recurrent Unit (GRU) corresponding to each visit. This introduces a second layer of temporal analysis beyond the inherent time-series nature of text and biomarker data.

\subsection{Multi-Task Learning Formulation in All Three Architectures}

\label{sec:mtl-all-models}

For all three pipelines, the final embeddings generated by the model pass through three separate heads responsible for one task each. The gradients between heads are not detached, allowing decisions from one task to influence the others and thus enabling us to leverage comorbidity (the effect of this is evaluated in Section \ref{sec:mtl-results}). Since the labels for each task can suffer from class imbalance, a weighted binary cross-entropy loss is used. The loss for a single task \(t\) is given by
\begin{equation}
    \mathcal{L}_t = -\left( w^+_t y_t \log(\hat{y}_t) + (1 - y_t) \log(1 - \hat{y}_t) \right)
\end{equation}
where \(y_t\) is the true label (0 or 1), \(\hat{y}_t\) is the predicted probability, and \(w^+_t\) is the weight applied to positive samples to account for class imbalance in a particular task.

For our configuration of one main task (\( M \)) and two auxiliary tasks (\(A_0, A_1\)), the total loss is
\begin{equation}
    \mathcal{L}_{\text{total}} = \mathcal{L}_M + \lambda_{\text{aux}} \left( \mathcal{L}_{A_0} + \mathcal{L}_{A_1} \right)
\end{equation}

where \(\mathcal{L}_M\) is the main task loss, \(\mathcal{L}_{A_0}\) and \(\mathcal{L}_{A_1}\) are auxiliary task losses, and \(\lambda_{\text{aux}}\) controls the weight of auxiliary losses.

\section{Results and Discussion}

Table \ref{tab:main-results} presents the classification performance across all three binary prediction tasks, categorized based on the subset of modalities utilized, the corresponding model architectures, and the base LLM employed. To ensure consistency, $\lambda_{\rm{aux}} = 0.25$ is used for all trials. While precision and recall are reported for completeness, balanced accuracy serves as our primary metric since it best captures aggregate performance across both positive and negatives cases.

To ensure a fair study, a Receiver Operating Characteristic (ROC) curve is constructed for each case based on the validation set, and thresholds are selected based on the point on the curve closest to the top left corner at $(0, 1)$. This threshold is then blindly applied to the test set and the resulting metrics are reported to ensure no information leakage between sets.

\subsection{Effect of Modality and Base LLM on Multimodal Prediction}
Table \ref{tab:main-results} shows an improvement in balanced accuracy as more modalities are incorporated into the Mental-Alpaca-based analysis, highlighting the utility of multimodal approaches. We also observe a significant improvement when using Mental-Alpaca as the base LLM, compared to LLaMA-2-7B, which agrees with the literature \cite{Xu_2024}. While LLaMA-2-7B has more weights, Mental-Alpaca is pre-trained on mental health tasks, leading to improved predictive power. The highest balanced accuracy across all three tasks (depression, suicidal ideation, and sleep disturbances) is observed when training the Mental-Alpaca model in the full trimodal regime.


\subsection{Utility of Longitudinal Analysis}
Table \ref{tab:main-results} demonstrates that the longitudinal enhancement to the full trimodal pipeline results in improved balanced accuracy for all three tasks in the case of LLaMA-2-7B, and to two tasks in the case of Mental-Alpaca. Overall, these results demonstrate that monitoring long-term subject trajectories leads to improved predictive power, which aligns with previous work on unimodal text datasets \cite{Qin_2023_LLMSocialMedia}.

\subsection{Effect of Multi-Task Learning} \label{sec:mtl-results}
For this experiment, the trimodal Mental-Alpaca-based longitudinal architecture is used, since it was found to perform best in the previous study. Table \ref{tab:aux-task-results} demonstrates the effect of increasing the importance (weightage) applied to the two auxiliary tasks on the ability to predict the main task as illustrated in (2). Increasing the weights on the auxiliary tasks leads to improved balanced accuracies for the main task, with the exception of $\lambda_{\rm{aux}} = 0.75$. This implies a synergy between the three tasks that is better leveraged when we incentivize the model to collectively learn all tasks.
\begin{table}
    \centering
    \caption{Effect of Auxiliary Task Weights on Primary Depression Task}
    \label{tab:aux-task-results}  
    \begin{tabular}{|c|c@{\hspace{0.8cm}}c@{\hspace{0.8cm}}c|} 
        \hline 
        \multirow{2}{*}{\textbf{Auxiliary Tasks Weights}} & \multicolumn{3}{c|}{\textbf{Primary Task Metrics}} \\ 
        \cline{2-4}  
        & \textbf{P} & \textbf{R} & \textbf{BA} \\ 
        \hline
        0.00 & 0.354 & 0.580 & 0.589 \\ 
        \hline 
        0.25 & 0.438 & 0.420 & 0.608 \\ 
        \hline 
        0.50 & 0.427 & 0.700 & 0.672 \\ 
        \hline 
        0.75 & 0.468 & 0.580 & 0.665 \\ 
        \hline
        1.00 & 0.535 & 0.620 & \textbf{0.708} \\ 
        \hline
    \end{tabular}
\end{table}
\section{Conclusion}
Our results demonstrate that deploying a longitudinal LLM-based model on speech data---treated as a trimodal multimedia source---enhances performance in predicting multiple mental health outcomes. By leveraging speech digital phenotypes, our approach captures rich behavioural markers, leading to improved performance compared to baseline models \cite{Xu_2024} on our dataset. To further substantiate these findings, we plan to benchmark our pipeline against publicly available datasets that have been used to assess baseline models, including Mental-Alpaca and other LLMs, in the context of comparable tasks. This work can also be extended to explore other LLM architectures, such as Mental-FLAN-T5 and GPT. Given that LLM outputs are heavily influenced by input prompts, conducting a more in-depth analysis of different prompt strategies would provide valuable insights.



\bibliographystyle{IEEEtran}
\bibliography{main}

\end{document}